\documentclass[journal,twoside]{IEEEtran}
\usepackage[T1]{fontenc}
\usepackage[utf8]{inputenc}
\usepackage{amsmath}
\usepackage{amssymb,bm}
\usepackage{graphicx,microtype,booktabs,array,tabularx,siunitx}
\usepackage{cite,url}
\usepackage[colorlinks=true,citecolor=blue,linkcolor=blue,urlcolor=black]{hyperref}
\graphicspath{{assets/}{assets/bio/}}

\ifdefined\DeclareUnicodeCharacter
\DeclareUnicodeCharacter{2010}{-}
\DeclareUnicodeCharacter{2011}{-}
\DeclareUnicodeCharacter{2212}{\ensuremath{-}}
\DeclareUnicodeCharacter{2009}{\,}
\DeclareUnicodeCharacter{202F}{\,}
\fi
\title{LumiTac: Compact, High-Rate Six-Axis Optical Tactile Sensing for Human–Robot Interaction}
\author{Hao~Wu,~\IEEEmembership{Student Member,~IEEE},
Haotian~Guo,~\IEEEmembership{Student Member,~IEEE},
Yu~Feng,~\IEEEmembership{IEEE/ASME Member},
Yutong~Wang,
and Jianshu~Zhou,~\IEEEmembership{IEEE/ASME Member}%
\thanks{This work was partially supported by the National University of Singapore Start-up Grant (NUS Start-up Grant, FY2026). Hao Wu, Haotian Guo, and Yu Feng contributed equally to this work. (Corresponding author: Jianshu Zhou.)}%
\thanks{Hao Wu, Haotian Guo, Yu Feng, Yutong Wang, and Jianshu Zhou are with the Department of Mechanical Engineering, National University of Singapore, Singapore (e-mail: jianshuzhou@nus.edu.sg).}%
}

\begin{document}
\maketitle
\thispagestyle{headings}

\begin{abstract}
Integrating six-axis tactile feedback into robotic interfaces requires compact sensors with accurate load measurement and rapid online reconstruction. This paper presents LumiTac, a low-profile, camera-free optoelectronic tactile sensor combining spatially distributed illumination with spectral encoding. Five photoemitters and five RGBW photoreceivers are mounted on opposing circuit boards coupled through translucent silicone. Load-induced interlayer motion modulates the internal light field under compression, shear, bending, and torsion. The sensing head measures 12~mm in diameter and 4.75~mm in thickness, weighs less than 2~g, and has a component cost of approximately USD~7. A compact multilayer perceptron reconstructs three force and three moment components from 20 optical channels. Independent testing yielded coefficients of determination ($R^2$) of 0.9657--0.9941, with mean absolute errors of 0.054--0.190~N for forces and 0.304--1.419~mN$\cdot$m for moments. Online prediction achieved an average update rate of 303.57~Hz and a median software processing latency of 0.722~ms. Experiments involving combined loading, light weights, and rapid tapping evaluated load tracking, sensitivity to small forces, and transient response, respectively. Robotic demonstrations further validated its applicability to tactile-guided manipulation and physical human--robot interaction. This study paves the way for compact, high-rate optical tactile sensing in robotic manipulation and human--robot interaction.
\end{abstract}


\section{Introduction}

\IEEEPARstart{R}{obotic} manipulation requires continuous adjustment to contact forces during grasping, alignment, and physical interaction. To enable responsive and adaptive manipulation, embedded sensing~\cite{yang2024selfsensing} and intuitive control interfaces~\cite{zhou2018intuitive,zhou2019grasping} have been explored to enhance robotic perception and interaction. In particular, tactile sensing provides local contact feedback that external vision alone cannot readily supply~\cite{kappassov2015tactile,dahiya2010tactile}. Beyond normal and tangential forces, bending and torsional moments characterize rotational loading relevant to contact alignment and orientation control. Measuring the complete six-axis contact wrench therefore supports multidimensional interaction feedback. Integrating this capability into space-constrained robotic interfaces, such as fingertips~\cite{mao2025flexible6axis}, grippers~\cite{leslie2023optical6axis}, and wearable devices~\cite{battaglia2016thimblesense}, requires sensing structures with a small footprint, limited thickness, and low mass. Rapid load reconstruction is equally important for timely feedback in contact-rich tasks.

\begin{figure}[t]
\centering
\includegraphics[width=0.975\columnwidth]{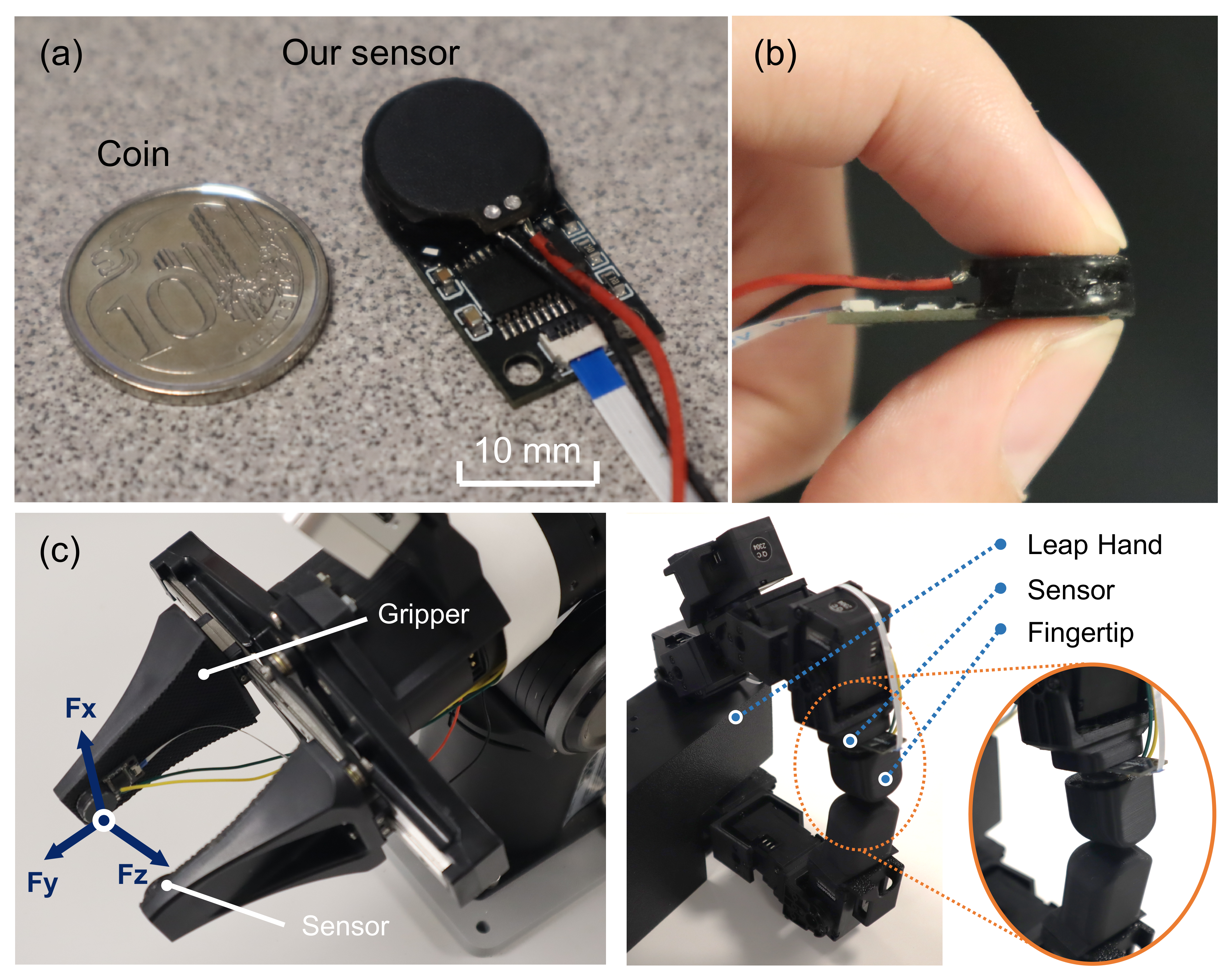}
\caption{LumiTac prototype and potential applications. (a) Size comparison with a coin; (b) side view of the sensor structure; (c) application concepts for robotic grippers and dexterous hands.}
\label{fig:prototype}
\end{figure}

\begin{table*}[!t]
\centering
\caption{Comparison of representative multiaxis tactile and force/torque sensors.}
\label{tab:compact_six_axis}
\begingroup
\footnotesize
\setlength{\tabcolsep}{3pt}
\renewcommand{\arraystretch}{1.12}
\newcommand{\ftcell}[1]{\begin{tabular}[t]{@{}l@{}}#1\end{tabular}}

\begin{tabularx}{\textwidth}{
@{}p{1.80cm}p{1.65cm}p{2.15cm}p{0.65cm}p{3.60cm}
>{\raggedright\arraybackslash}Xp{1.65cm}p{0.76cm}@{}
}
\toprule
Sensor & Transduction & Size [mm] &
\ftcell{Mass\\(g)} &
\ftcell{Reported load ranges\\$F$: N; $M$: mN$\cdot$m} &
\ftcell{Accuracy\\$F$: N; $M$: mN$\cdot$m} &
\ftcell{Sampling\\rate} & \ftcell{Cost\\(USD)} \\
\midrule

\ftcell{LiVec finger\\\cite{leslie2023optical6axis}} &
Optical &
\ftcell{$26.44\times38.22$\\$\times12$} &
NR &
\ftcell{$F$: $(\pm2,\pm2,0\text{--}12)$\\$M$: NR} &
\ftcell{Error SD\\
$F$: $(0.112,0.092,0.139)$\\
$M$: $(1.90,1.54,1.26)$} &
390 Hz & NR \\
\addlinespace[3pt]

\ftcell{El-Azizi et al.\\\cite{elazizi2026led}} &
Optical &
$\phi27\times20.82$ &
16.4 &
\ftcell{$F$: $\sim(\pm5,\pm5,0\text{--}5)$\\$M$: $\sim(\pm200,\pm200,\pm200)$} &
\ftcell{MAE\\
$F$: $(0.060,0.050,0.070)$\\
$M$: $(2.62,2.68,0.64)$} &
$>2.5$ kHz & 100 \\
\addlinespace[3pt]

\ftcell{F-TOUCH\\\cite{li2021ftouch}} &
Visuotactile &
NR &
NR &
\ftcell{$F$: $(\pm1,\pm1,0\text{--}8)$\\$M$: $(\pm20,\pm20,\pm20)$} &
\ftcell{RMSE\\
$F$: $(0.018,0.016,0.094)$\\
$M$: $(0.657,1.566,0.407)$} &
30 Hz & NR \\
\addlinespace[3pt]

\ftcell{Mao et al.\\\cite{mao2025flexible6axis}} &
Thermal &
$12\times15\times5$ &
0.30 &
\ftcell{$F$: $(\pm5,\pm5,0\text{--}15)$\\$M$: $(\pm60,\pm60,\pm20)$} &
\ftcell{RMSE\\
$F$: $(0.180,0.150,0.180)$\\
$M$: $(0.58,0.62,0.26)$} &
NR & NR \\
\addlinespace[3pt]

\ftcell{CoinFT\\\cite{choi2026coinft}} &
Capacitive &
$\phi20\times2$ &
2.0 &
\ftcell{$F$: $(\pm5,\pm5,0\text{--}15)$\\$M$: $(\pm100,\pm100,\pm80)$} &
\ftcell{RMSE\\
$F$: $(0.186,0.134,0.186)$\\
$M$: $(1.114,1.322,0.810)$} &
360 Hz & $<11$ \\
\addlinespace[3pt]

\ftcell{HexaTouch\\\cite{song2026hexatouch}} &
Capacitive &
$15\times15\times8$ &
NR &
\ftcell{$F$: $(\pm5,\pm5,0\text{--}25)$\\$M$: $(\pm100,\pm100,\pm50)$} &
\ftcell{RMSE\\
$F$: $(0.140,0.130,0.130)$\\
$M$: $(2.48,2.50,0.95)$} &
NR & NR \\
\midrule

\ftcell{\textbf{LumiTac}\\\textbf{(this work)}} &
Optical &
$\phi12\times4.75$ &
$<2$ &
\ftcell{$F$: $(\pm10,\pm10,0\text{--}30)$\\
$M$: $(\pm100,\pm100,\pm20)$} &
\ftcell{MAE\\
$F$: $(0.056,0.054,0.190)$\\
$M$: $(0.774,1.419,0.304)$} &
$\sim300$ Hz & $\sim7$ \\

\bottomrule
\end{tabularx}

\smallskip
\begin{minipage}{\textwidth}
\scriptsize
Components are ordered $x,y,z$. MAE: mean absolute error; RMSE: root-mean-square error; NR: not reported; SD: standard deviation.
\end{minipage}

\endgroup
\end{table*}

Multi-axis force/torque (F/T) measurement has been implemented using various transduction mechanisms, including capacitive~\cite{lee2016capacitive,choi2026coinft}, magnetic~\cite{lu2023magtac}, thermal~\cite{mao2025flexible6axis}, piezoelectric~\cite{gong2026piezoceramic}, and optical~\cite{wang2026flexiray,elazizi2026led} sensing. Capacitive designs organize deformation-dependent responses through electrode geometry and spacing \cite{lee2016capacitive}. Lee et al. improved directional sensitivity through isotropic optimization and redundant measurements \cite{lee2025isotropic}. CoinFT  achieves a thin, low-cost structure using silicone pillars and configurable electrode readout \cite{choi2026coinft}, while HexaTouch reconstructs six-axis loads and contact position from capacitive images \cite{song2026hexatouch}. These studies demonstrate the potential of capacitive sensing for compact six-axis measurement. However, parasitic capacitance and electrode misalignment can introduce measurement errors, while cross-axis coupling complicates calibration and the reconstruction of individual force and moment components.

Magnetic tactile sensors infer contact loads from deformation-induced field changes detected by Hall elements \cite{lu2023magtac}. MagCilia combines flexible magnetic cilia with a Hall sensor to reconstruct normal and tangential forces from field changes induced by compression and bending \cite{feng2026magcilia}. Veliky et al. use distributed Hall-sensor and magnet pairs to capture translations and rotations of a silicone-supported structure for six-axis load reconstruction \cite{veliky2024soft6axis}. Although magnetic sensing enables compliant multiaxis measurement, external magnetic fields and nearby ferromagnetic objects can distort the measured field, introducing errors that require shielding or compensation.

Thermal and piezoelectric transduction offer further approaches to six-axis sensing. Mao et al. used thin-film thermistors to detect deformation-induced changes in the thermal conductance of a compliant medium, enabling an ultralight design but requiring active heating and temperature compensation \cite{mao2025flexible6axis}. Gong et al. combined an ultrathin piezoceramic shell with patterned electrodes to distinguish load-induced charge distributions. This monolithic architecture supports sensitive dynamic measurements, although its AC-coupled readout limits absolute quasi-static load measurement \cite{gong2026piezoceramic}.

Vision-based tactile sensors encode contact-induced deformation in camera images and infer forces from surface appearance or marker displacement, as demonstrated by GelSight, TacTip, and DIGIT Pinki \cite{yuan2017gelsight,wardcherrier2018tactip,di2025pinki}. Camera-based pose tracking enables six-axis force and moment reconstruction in fiducial-based sensors \cite{fernandez2021visiflex}. Optical feedback has been integrated into compliant grippers for manipulation and force regulation \cite{lu2025dexitac,wang2026flexiray,wang2026relaxation}. Despite their high spatial resolution, camera acquisition and image processing can limit update rates. Camera size, optical working distance, and computational demands also complicate integration into thin, compact robotic interfaces.

Camera-free optical sensors recover loads from wavelength shifts or discrete light measurements. Fiber Bragg grating-based designs distribute strain measurements along an optical fiber to achieve low-coupling six-axis sensing, although they require fiber attachment and wavelength interrogation \cite{xiong2021fbg6axis}. Discrete optoelectronic designs instead measure deformation-dependent light signals using distributed emitters and receivers \cite{demaria2012force,palli2014optoelectronic}. LiVec combines local optical measurements to recover forces, displacements, and resultant moments \cite{leslie2023optical6axis}, whereas El-Azizi et al. use LEDs as both emitters and receivers to measure motion between elastomer-connected plates \cite{elazizi2026led}. By avoiding image formation, discrete optical readout supports compact layouts, low-dimensional signal processing, and high-rate acquisition. Ambient light, optical alignment, and elastomer nonlinearity nevertheless remain design and calibration considerations.

A central challenge in compact six-axis optical sensing is to distinguish coupled translations and rotations, since different deformation modes may produce similar changes in received light intensity. Previous studies have explored overlapping light signals and patterned color encoding to introduce spatial and spectral information into deformation measurements \cite{piacenza2020overlap,yu2026patterned}. The comparison in Table~\ref{tab:compact_six_axis} further motivates combining informative optical measurements with a low-profile structure and rapid online reconstruction. To address these requirements, we present LumiTac, a camera-free optoelectronic tactile sensor shown in Fig.~\ref{fig:prototype}. Five photoemitters and five RGBW photoreceivers are mounted on opposing PCBs coupled through translucent silicone. The spatially offset emitter--receiver layout and multiple LED colors provide spatial and spectral cues associated with load-induced interlayer motion. A compact multilayer perceptron (MLP) jointly processes the resulting 20 optical channels to estimate three force and three moment components, enabling frame-wise online load reconstruction. The main contributions are as follows:

\begin{itemize}
\item A miniaturized hardware design that integrates six-axis tactile sensing into a head measuring 12~mm in diameter and 4.75~mm in thickness, with a mass below 2~g and an estimated component cost of approximately USD~7.
\item An efficient six-axis reconstruction pipeline achieving mean absolute errors (MAEs) of 0.054--0.190~N for forces and 0.304--1.419~mN$\cdot$m for moments, with coefficients of determination ($R^2$) of 0.9657--0.9941 on an independent test recording. Its online implementation delivers an average prediction update rate of 303.57~Hz with a median software processing latency of 0.722~ms.
\item Experimental validation of load tracking, sensitivity to small forces, and transient response, together with demonstrations of simulated manipulator guidance and physical robot admittance control, establishing a single tactile interface for robot motion and gripper operation.
\end{itemize}

\section{Sensor Design}

\subsection{Layered Structure}

\begin{figure}[t]
\centering
\includegraphics[width=0.96\columnwidth]{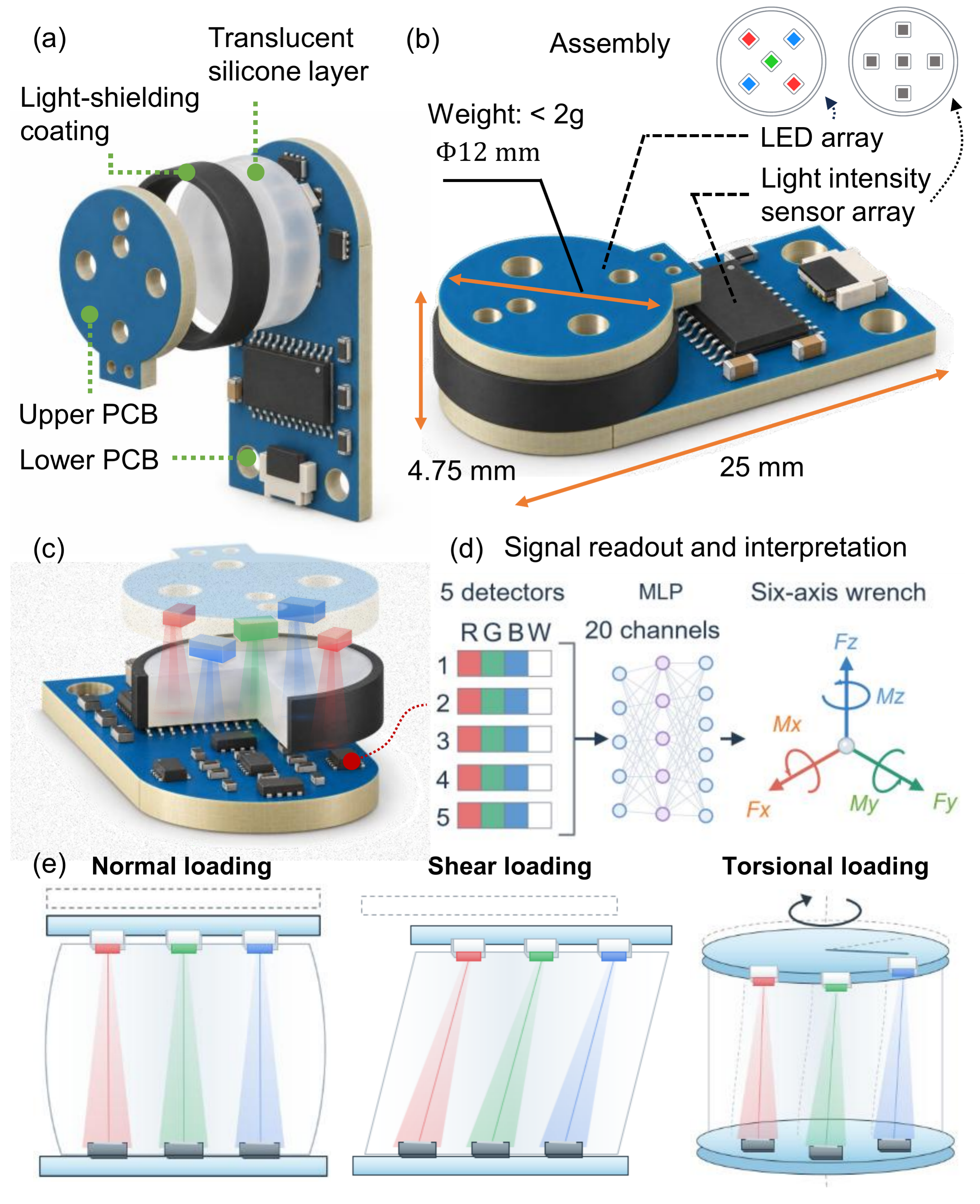}
\caption{Structure and sensing principle of LumiTac. (a) Layered structure of the sensor; (b) assembled sensor and arrangement of light-emitting elements and receivers; (c) cross-sectional view of the internal optical paths; (d) 20-channel RGBW signal readout and MLP-based six-axis force and moment reconstruction; (e) schematic illustration of optical path changes under normal, shear, and torsional loading.}
\label{fig:structure}
\end{figure}

LumiTac employs a layered optical architecture comprising two opposing PCBs connected by a translucent silicone layer and enclosed by a peripheral light-blocking coating, as shown in Fig.~\ref{fig:structure}(a). The upper PCB receives external loads and carries the photoemitters, while the fixed lower PCB accommodates the photoreceivers and readout circuitry. The silicone layer serves as both a compliant mechanical connection and an optical propagation medium. Under external loading, its deformation induces small relative translations and rotations between the emitting and receiving planes, altering the internal light field and the intensities measured by the photoreceivers. The surrounding coating suppresses ambient-light interference, providing a more stable environment for internal light propagation.

As shown in Fig.~\ref{fig:structure}(b), the circular sensing head measures 12~mm in diameter and 4.75~mm in thickness, with a mass below 2~g. The evaluated load ranges are $\pm10$~N for $F_x$ and $F_y$, 0--30~N for the compressive magnitude $|F_z|$, $\pm100$~mN$\cdot$m for $M_x$ and $M_y$, and $\pm20$~mN$\cdot$m for $M_z$. The lower PCB incorporates I$^2$C multiplexing to read the five photoreceivers and provides the electrical interface, extending the overall board length to 25~mm. With an estimated component cost of approximately USD~7 per unit, this compact and lightweight architecture facilitates affordable integration of six-axis tactile sensing into robotic fingertips and grippers.

\begin{figure*}[t]
\centering
\includegraphics[width=0.915\textwidth]{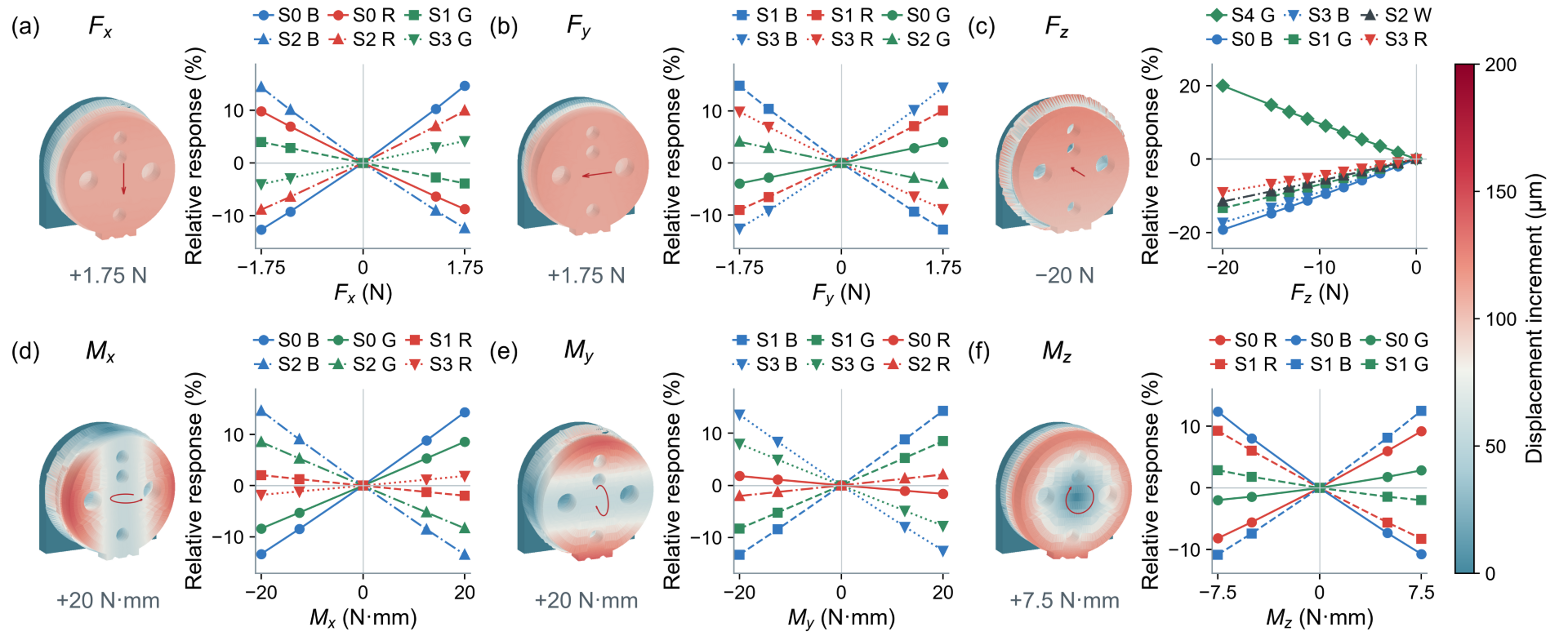}
\caption{Simulated deformation and optical responses of LumiTac under single-axis loading. (a)--(c) Responses to $F_x$, $F_y$, and $F_z$, respectively; (d)--(f) responses to $M_x$, $M_y$, and $M_z$, respectively. Each panel shows selected representative RGBW responses and deformation maps colored by displacement increment.}
\label{fig:sim}
\end{figure*}

\subsection{Working Principle}

External loading deforms the silicone layer and changes the relative positions and orientations of the emitters and receivers. Normal compression changes the separation between the two boards, shear produces lateral displacement, bending tilts the emitting plane, and torsion rotates it relative to the receiving plane. These movements alter the optical path lengths, illumination directions, and spectral contributions measured at different receiver locations, as illustrated in Fig.~\ref{fig:structure}(e).

Each receiver measures red, green, blue, and broadband clear intensities, denoted R, G, B, and W, respectively. The outputs of the five receivers form a 20-dimensional optical observation:
\begin{equation}
\begin{aligned}
\mathbf{I}_i
&=[I_{iR},I_{iG},I_{iB},I_{iW}]^{\mathsf{T}},\\
\mathbf{I}
&=[\mathbf{I}_0^{\mathsf{T}},\ldots,
   \mathbf{I}_4^{\mathsf{T}}]^{\mathsf{T}}
   \in\mathbb{R}^{20}.
\end{aligned}
\label{eq:optical_observation}
\end{equation}

Deformation changes the RGBW intensities measured at the five photoreceiver locations, with compression, shear, bending, and torsion producing different patterns of intensity changes. As illustrated in Fig.~\ref{fig:structure}(d), the RGBW readings from the five receivers form a normalized 20-dimensional MLP input. The network jointly maps these spatial and spectral responses to three force components ($F_x,F_y,F_z$) and three moment components ($M_x,M_y,M_z$), including under combined loading.

\subsection{Emitter--Receiver Arrangement and Design Rationale}

The emitting plane contains a central green LED and four peripheral LEDs with alternating red and blue colors. The receiving plane contains four peripheral receivers, S0--S3, and a central receiver, S4. As illustrated in Fig.~\ref{fig:structure}, the peripheral emitter and receiver patterns are angularly offset, forming multiple oblique optical paths between the two planes.

The resulting responses were examined by combining finite-element deformation analysis with a simplified optical model. In Fig.~\ref{fig:sim}(a) and (b), $F_x$ produces pronounced red and blue channel changes at S0 and S2, whereas $F_y$ produces corresponding changes at S1 and S3. Reversing the force direction reverses these response trends. The receiver locations therefore provide spatial information about the applied loading direction. Under normal compression, the central green channel, S4-G, increases while the selected peripheral channels decrease, as shown in Fig.~\ref{fig:sim}(c). These responses show that measurements from the central and peripheral receivers provide complementary information about load-induced deformation.

The moment responses also depend on receiver location and optical channel. For $M_x$ and $M_y$, the blue and green responses vary across S0/S2 and S1/S3, respectively, as shown in Fig.~\ref{fig:sim}(d) and (e). Under $M_z$, the red and blue channels change in opposite directions within each of the illustrated peripheral receivers, S0 and S1, as shown in Fig.~\ref{fig:sim}(f). Each independently applied force or moment component thus produces a characteristic spatial and spectral response pattern. These simulated response patterns are consistent with the channel response trends observed in the physical sensor, providing an intuitive physical basis for six-axis load decoupling using a compact multichannel model.

\section{Calibration and Signal Interpretation}

\subsection{Reference Measurements and Preprocessing}

\begin{figure*}[t]
\centering
\includegraphics[width=0.875\textwidth]{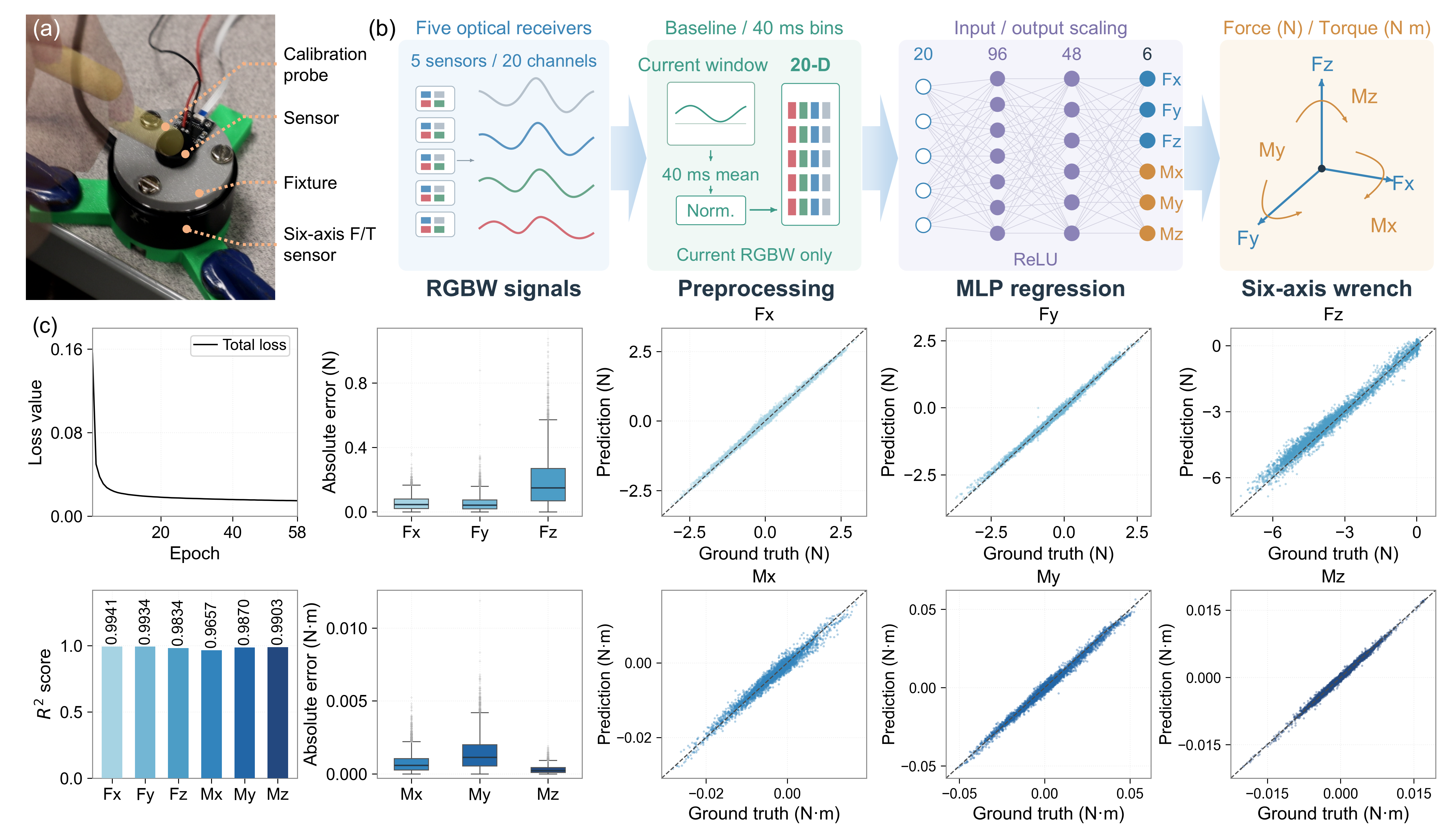}
\caption{Six-axis load calibration and reconstruction.
(a) Calibration setup with a reference F/T sensor.
(b) 20-channel RGBW preprocessing and MLP-based reconstruction.
(c) Training loss and independent test results, including per-axis $R^2$, absolute error distributions, and
predicted--reference comparisons.}
\label{fig:network}
\end{figure*}

A six-axis F/T sensor provided reference measurements for calibrating the tactile sensor. As shown in Fig.~\ref{fig:network}(a), the tactile sensor was mounted on the reference F/T sensor, and loads were applied to its sensing surface using a probe. The 20 optical channels from the five RGBW receivers were recorded alongside the reference six-axis measurements. The host computer logged each optical sample with the latest available reference measurement.

For each recording, the baseline $b_j$ was defined as the median reading of optical channel $j$ during the initial 2~s unloaded period. The optical measurements and reference loads were then averaged over nonoverlapping 40~ms windows, and the optical response was normalized as
\begin{equation}
r_{j,k}
=
\frac{\bar{I}_{j,k}-b_j}
{\max\left(|b_j|,\epsilon\right)},
\qquad j=1,\ldots,20,
\label{eq:baseline_response}
\end{equation}
where $\bar{I}_{j,k}$ is the mean reading of channel $j$ in window $k$, and $\epsilon>0$ prevents division by zero. This normalization expresses the optical response relative to the unloaded state and reduces differences in baseline magnitude across channels.

\subsection{Compact Six-Output Regression Model}

The compact MLP was used to reconstruct the three force and three moment components from the 20-channel optical response. The network consists of two hidden layers with 96 and 48 neurons, respectively, each followed by rectified linear unit (ReLU) activation, and a six-dimensional linear output layer. The reconstruction is expressed as
\begin{equation}
\begin{aligned}
\hat{\mathbf{w}}
&=
\mathcal{S}_{w}^{-1}
\left[f_{\theta}\!\left(\mathcal{S}_{r}(\mathbf{r})\right)\right], \\
\mathbf{w}
&=[F_x,F_y,F_z,M_x,M_y,M_z]^{\mathsf{T}},
\end{aligned}
\label{eq:load_reconstruction}
\end{equation}
where $f_{\theta}$ denotes the MLP, and $\mathcal{S}_{r}$ and $\mathcal{S}_{w}$ standardize the optical inputs and reference loads component-wise, respectively. 

Standardizing the target loads accounts for the different units and numerical scales of the six output components during training. Fig.~\ref{fig:network}(c) summarizes the training and independent test results obtained using the 40~ms offline preprocessing. The network contains 6,966 trainable parameters and requires approximately 6,816 weight multiplications per prediction, keeping the computational cost of load reconstruction low.

For online operation, the trained network and fitted standardization parameters were reused. The optical signals were processed using a causal moving average with a nominal 20~ms window, updated after each valid optical frame. The averaged signals were then baseline-normalized and processed according to~\eqref{eq:load_reconstruction}, producing a new load estimate after each valid optical frame.

The online implementation achieved an average prediction update rate of 303.57~Hz, with a median software processing latency of 0.722~ms. This latency measures host-side processing and excludes sensor acquisition, data transmission, and display. These results demonstrate that the compact model supports high-rate load reconstruction with predictions updated at the incoming optical frame rate.

\subsection{Six-Axis Reconstruction Accuracy}

The MLP model maps the 20-channel optical intensity measurements to three force and three moment components. Reconstruction performance was evaluated using an independent test recording excluded from training. As shown in Fig.~\ref{fig:network}(c), the $R^2$ values for $F_x$, $F_y$, and $F_z$ are 0.9941, 0.9934, and 0.9834, respectively, while those for $M_x$, $M_y$, and $M_z$ are 0.9657, 0.9870, and 0.9903, respectively, yielding a mean $R^2$ of 0.9857 across the six components. The predicted--reference pairs lie close to the identity line, indicating close agreement with the reference measurements.

The MAEs for $F_x$, $F_y$, and $F_z$ are 0.056, 0.054, and 0.190~N, respectively, while those for $M_x$, $M_y$, and $M_z$ are 0.774, 1.419, and 0.304~mN$\cdot$m, respectively. The corresponding absolute error distributions are presented in Fig.~\ref{fig:network}(c). These results demonstrate accurate reconstruction of all six load components from the shared optical measurements.

\section{Sensor Performance Evaluation}

\subsection{Real-Time Response under Combined Loading}

\begin{figure*}[t]
\centering
\includegraphics[width=0.88\textwidth]{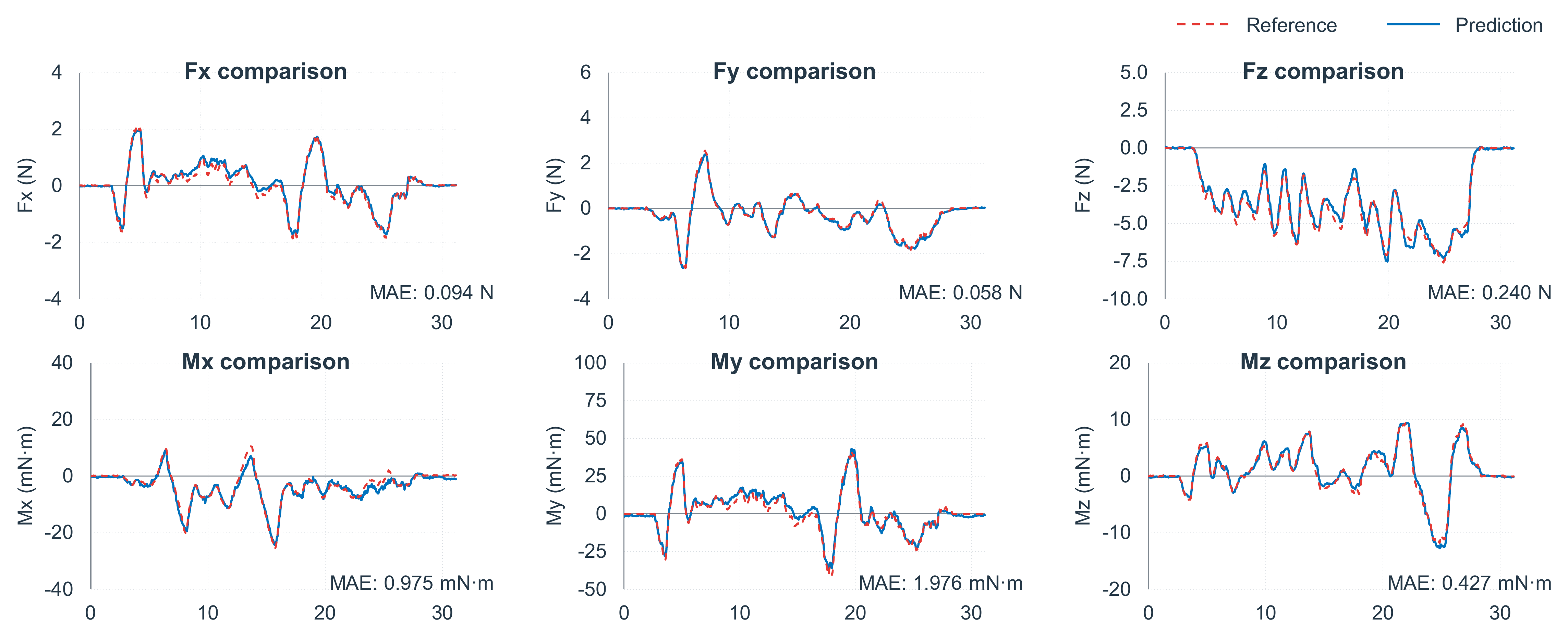}
\caption{Real-time six-axis force and moment reconstruction under combined loading. Reference measurements are shown as red dashed lines and smoothed predictions as blue solid lines, with the mean absolute error (MAE) annotated for each axis. Time is in seconds.}
\label{fig:realtime}
\end{figure*}

Real-time six-axis load estimation was evaluated by applying continuously varying loads to the sensing surface. To examine diverse contact conditions, the loading sequence combined normal compression, shear forces, and bending and torsional moments, with varying amplitudes and direction reversals. The trained model processed the multichannel optical measurements, and its predictions were compared with the paired reference force and moment measurements. The records in Figs.~\ref{fig:realtime} and~\ref{fig:evaluation} were processed at 25~Hz; online timing was evaluated separately.

As shown in Fig.~\ref{fig:realtime}, the filtered predictions closely track the reference signals, capturing the principal changes in both load magnitude and direction. Over the approximately 30~s experiment, the MAEs for $F_x$, $F_y$, and $F_z$ are 0.094, 0.058, and 0.240~N, respectively, while those for $M_x$, $M_y$, and $M_z$ are 0.975, 1.976, and 0.427~mN$\cdot$m, respectively. The supplementary video complements these quantitative results by showing continuous six-axis estimation during repeated contacts, combined multi-axis loading, and contacts applied at different rates, illustrating the sensor's responses as the contact conditions change. These results demonstrate real-time reconstruction of time-varying contact forces and moments, supporting continuous six-axis feedback for robotic interaction and manipulation.

\subsection{Sensitivity and Transient Response}

\begin{figure}[t]
\centering
\includegraphics[width=0.96\columnwidth]{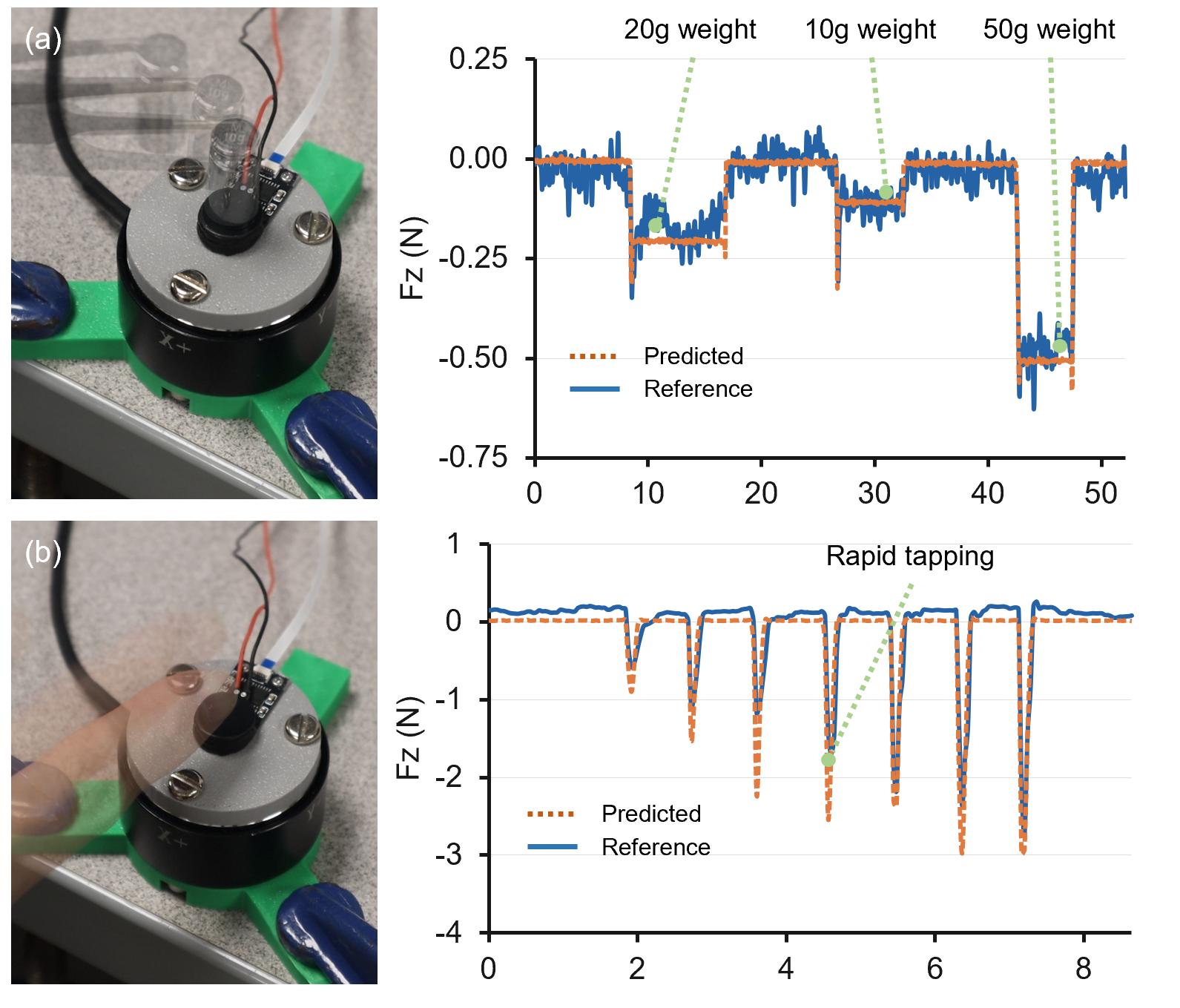}
\caption{Normal force responses under weight loading and rapid tapping. (a) Concatenated records of loading with 20~g, 10~g, and 50~g weights; (b) rapid finger tapping. Time is in seconds.}
\label{fig:evaluation}
\end{figure}

The sensor's response to small sustained loads was examined using standard weights of 20, 10, and 50~g, which were individually placed on and removed from the sensing surface. The estimated normal force $F_z$ was recorded alongside the reference measurements. As shown in Fig.~\ref{fig:evaluation}(a), the predictions follow the reference load levels and form distinct plateaus during the holding periods, returning toward the unloaded baseline after weight removal. A clear response is observed even for the 10~g weight, corresponding to approximately 0.1~N. The agreement across the three load levels demonstrates the sensor's ability to detect small compressive loads and track sustained contact forces.

To assess the sensor's response to transient contact loads, rapid finger taps were applied to the sensing surface, producing successive loading and unloading events. As shown in Fig.~\ref{fig:evaluation}(b), the estimated normal force captures all six taps, with peak compressive force magnitudes ranging from approximately 0.9 to 3.1~N. Comparison with the reference measurements shows that the predictions track the principal force variations during contact and release, capturing changes in peak amplitude and the return toward the unloaded baseline between taps. These results demonstrate the sensor's ability to resolve successive transient contacts and track rapidly varying normal loads over the tested force range.

\section{Robotic Applications}

\subsection{Handheld Control of a Simulated Robotic Manipulator}

\begin{figure*}[t]
\centering
\includegraphics[width=0.925\textwidth]{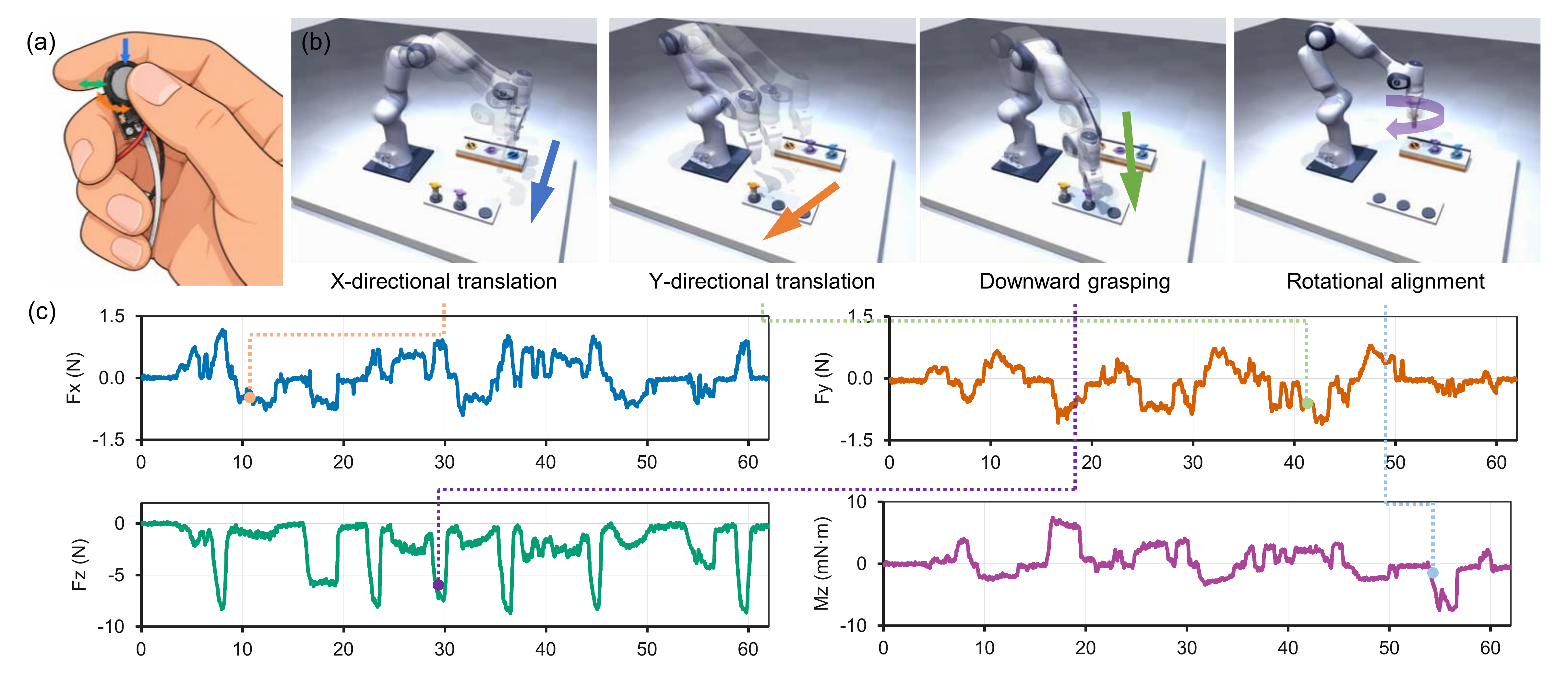}
\caption{Interactive manipulation of a simulated Franka robotic arm using LumiTac sensor inputs. (a) Handheld sensor input; (b) planar translation, downward grasping, and rotational alignment operations; (c) corresponding $F_x$, $F_y$, $F_z$, and $M_z$ control inputs. Time is in seconds.}
\label{fig:interaction}
\end{figure*}

The compact form factor and multi-axis force and moment sensing capability of LumiTac make it suitable for use as a portable interface for human--robot interaction. In this demonstration, an operator controlled a simulated Franka robotic arm by applying loads to the handheld sensor, as illustrated in Fig.~\ref{fig:interaction}(a) and (b). The shear forces $F_x$ and $F_y$ commanded planar translation along two orthogonal directions, while the normal force $F_z$ provided input for downward motion and grasping and placement operations. The torsional moment $M_z$ controlled end-effector rotation to align the grasped object with its target.

During the manipulation task, the operator adjusted the magnitude and direction of the applied loads according to the observed arm pose and target location, guiding the arm through object approach, grasping, alignment, and placement. Fig.~\ref{fig:interaction}(c) records the four control inputs over a sequence lasting approximately 60~s. The $F_x$, $F_y$, and $M_z$ traces show inputs of both signs for translational and rotational adjustments, whereas $F_z$ exhibits successive compressive inputs associated with downward motion and object handling. The signal trajectories show how the operator adjusted individual control inputs and coordinated them to produce combined robot motions during the task.

\subsection{Tactile-Guided Admittance Interaction}

\begin{figure*}[t]
\centering
\includegraphics[width=0.88\textwidth]{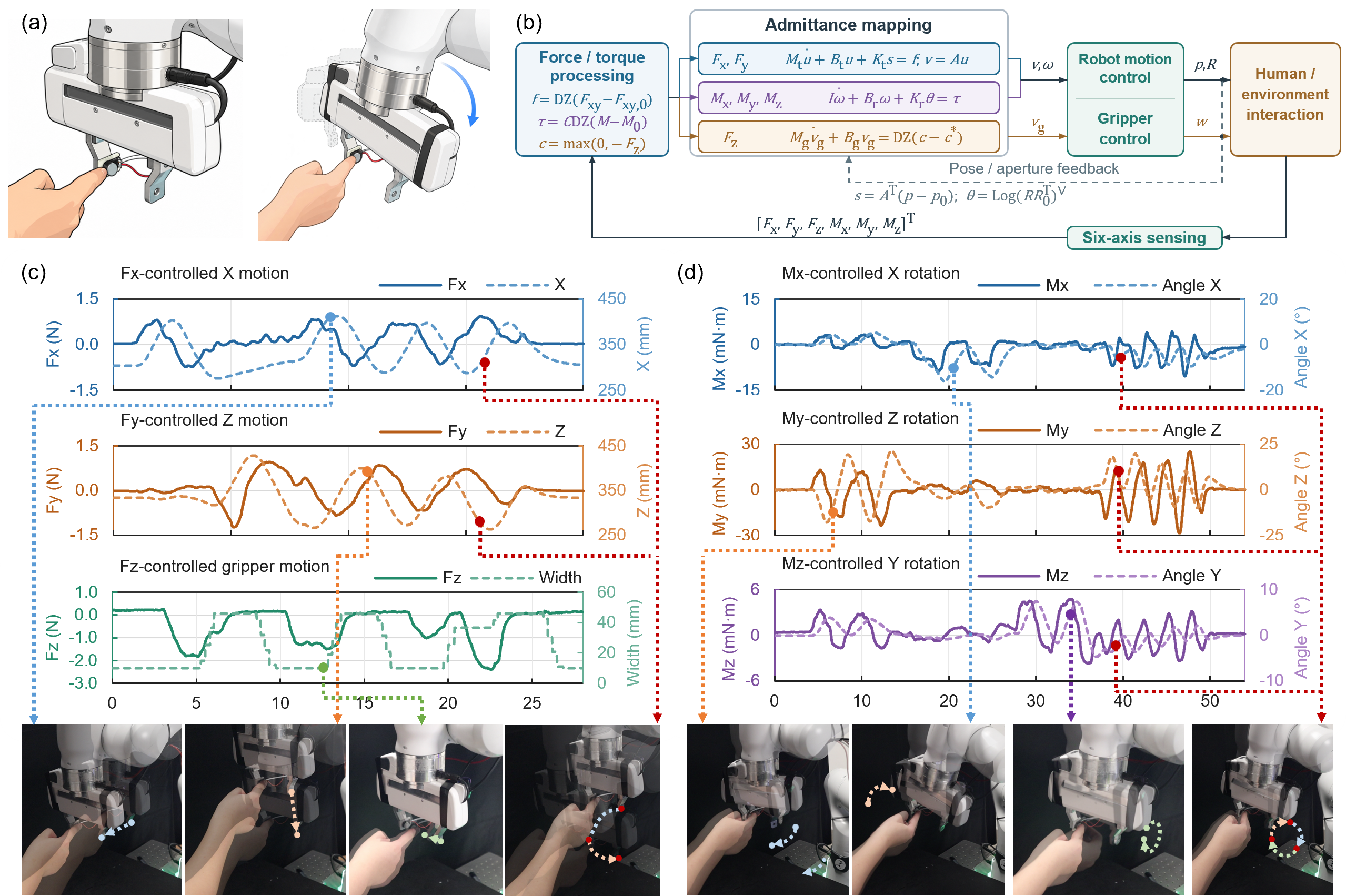}
\caption{Human--robot interaction experiments using LumiTac. (a) Integration of LumiTac into the robotic gripper and manual load application; (b) admittance-based control framework; (c) translational motion and gripper aperture responses; (d) rotational motion responses. Time is in seconds.}
\label{fig:robot}
\end{figure*}

Beyond handheld operation, LumiTac can be further integrated into a robot end effector to provide a direct tactile interface for guiding robot motion and gripper operation. As shown in Fig.~\ref{fig:robot}(a), LumiTac was mounted on the robot gripper, allowing the operator to apply pushing, pressing, and twisting inputs directly to its sensing surface. The reconstructed forces and moments were mapped to end-effector translation, gripper aperture, and orientation.

Fig.~\ref{fig:robot}(b) illustrates the control framework. After bias compensation and dead-zone processing, the measured forces and moments were converted into motion references through virtual admittance models. Specifically, $F_x$ and $F_y$ controlled translation along the robot $X$ and $Z$ directions, while $F_z$ controlled gripper opening and closing. The moment components $M_x$, $M_y$, and $M_z$ controlled rotation about the robot $X$, $Z$, and $Y$ axes, respectively.

The translational and rotational responses were governed by
\begin{equation}
\begin{aligned}
\mathbf{M}_t\ddot{\mathbf{s}}
+\mathbf{B}_t\dot{\mathbf{s}}
+\mathbf{K}_t\mathbf{s}
&= \mathbf{f}_c, \\
\mathbf{I}_r\ddot{\boldsymbol{\theta}}
+\mathbf{B}_r\dot{\boldsymbol{\theta}}
+\mathbf{K}_r\boldsymbol{\theta}
&= \boldsymbol{\tau}_c,
\end{aligned}
\label{eq:tactile_admittance}
\end{equation}
where $\mathbf{s}=[\Delta X,\Delta Z]^{\mathsf{T}}$ is the translational displacement from the selected reference position, and $\boldsymbol{\theta}$ denotes the local rotational displacement. The matrices $\mathbf{M}_t$ and $\mathbf{I}_r$ define virtual translational and rotational inertia, respectively, while $\mathbf{B}_t$ and $\mathbf{B}_r$ define damping, and $\mathbf{K}_t$ and $\mathbf{K}_r$ define stiffness. The processed inputs $\mathbf{f}_c$ and $\boldsymbol{\tau}_c$ incorporate the gains and direction mappings applied to $[F_x,F_y]^{\mathsf{T}}$ and $[M_x,M_z,M_y]^{\mathsf{T}}$, respectively. These virtual dynamics shape the motion response to the operator's applied loads. A separate aperture branch used the compressive input $c=\max(0,-F_z)$, together with a selected threshold, to generate gripper commands. Motion velocities and gripper aperture were bounded by their configured limits.

Fig.~\ref{fig:robot}(c) and (d) show the measured forces and moments alongside the corresponding robot motion. The translational, rotational, and gripper responses follow changes in the applied loads through repeated manual adjustments, showing consistent correspondence between tactile inputs and the resulting motion. By adapting its motion to the operator's applied forces and moments, the robot supported compliant adjustment of its position and orientation, enabling direct physical guidance of the end effector. These results demonstrate that LumiTac provides a direct tactile interface for guiding translation, orientation, and gripper operation, extending its use from contact-load measurement to physical human--robot interaction.

\section{Conclusion}
This paper presented LumiTac, a low-profile optoelectronic tactile sensor that combines spatially distributed illumination and spectral encoding for six-axis load measurement. Opposing photoemitter and photoreceiver arrays, coupled through a translucent silicone layer, convert load-induced interlayer motion into 20-channel optical responses, from which a compact MLP jointly reconstructs three force and three moment components. The sensing head measures 12~mm in diameter and 4.75~mm in thickness, with an estimated component cost of approximately USD~7 per sensor. These results demonstrate the integration of compact mechanical design, multichannel optical sensing, and efficient six-axis load reconstruction in an affordable tactile sensor.

Beyond measurement, LumiTac served as a tactile input interface for both a simulated robotic arm and a physical robotic gripper. The reconstructed forces and moments supported translational guidance, orientation adjustment, and gripper operation through handheld input and admittance control. These demonstrations connect compact six-axis contact sensing with tactile-guided robot motion. Future work will explore integration into dexterous hands, wearable interfaces, and the feet of legged robots to support broader applications in robotic manipulation, human--robot interaction, and foot--ground interaction sensing.
\bibliographystyle{IEEEtran}
\bibliography{references}

\begin{IEEEbiography}[{\includegraphics[width=1in,height=1.25in,clip,keepaspectratio]{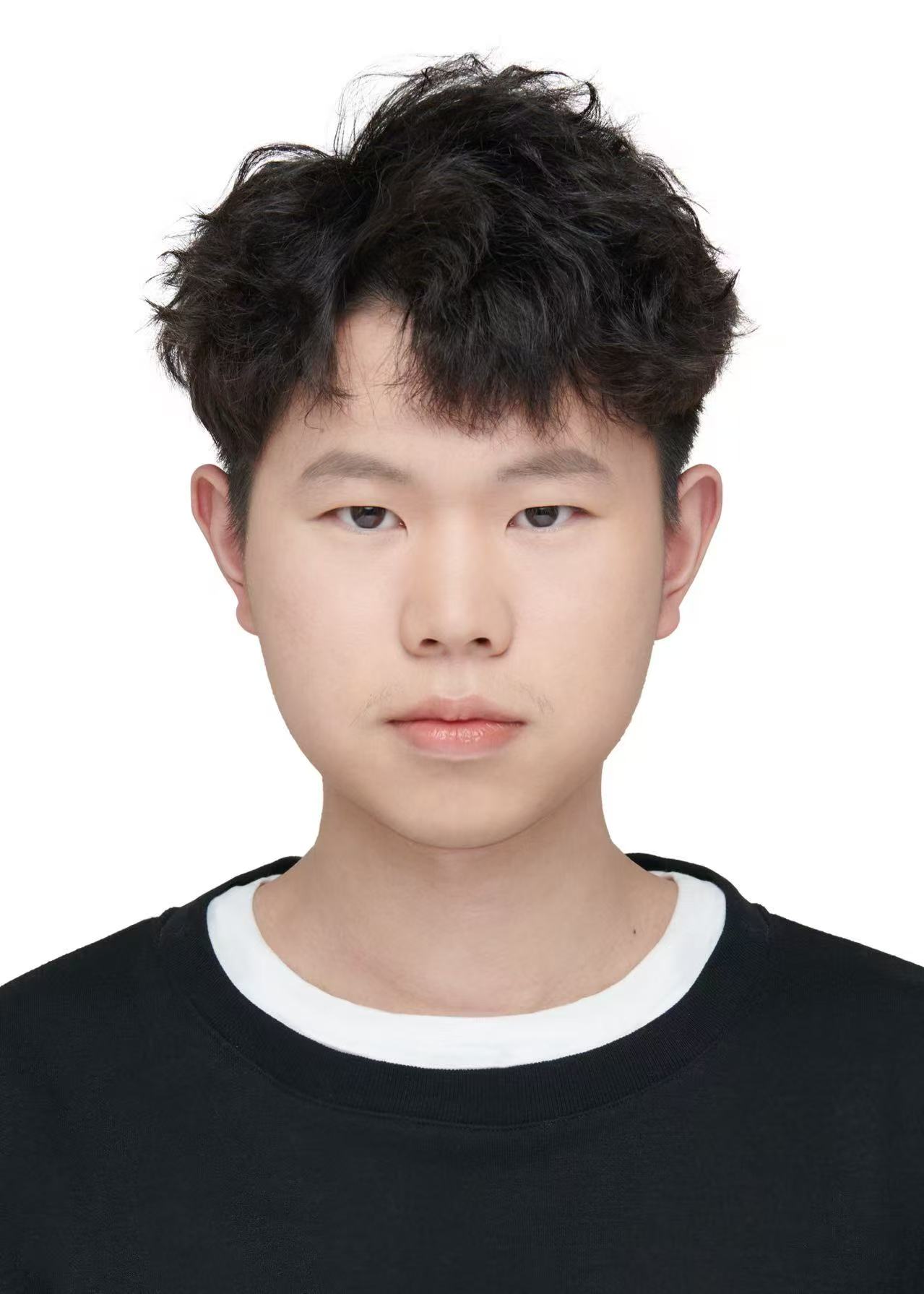}}]{Hao Wu}
(Student Member, IEEE) received the B.Eng. degree in mechanical engineering from Zhejiang University, China, in 2024. From 2024 to 2026, he was a Research Assistant with the Robotic Perception and Grasp Lab at Zhejiang University, China. He is currently pursuing the Ph.D. degree in mechanical engineering at the National University of Singapore, Singapore, where he is a member of the Robotics and Manipulation Lab. His research interests include mechanism design, soft robotics, and multimodal tactile sensing.\end{IEEEbiography}

\begin{IEEEbiography}[{\includegraphics[width=1in,height=1.25in,clip,keepaspectratio]{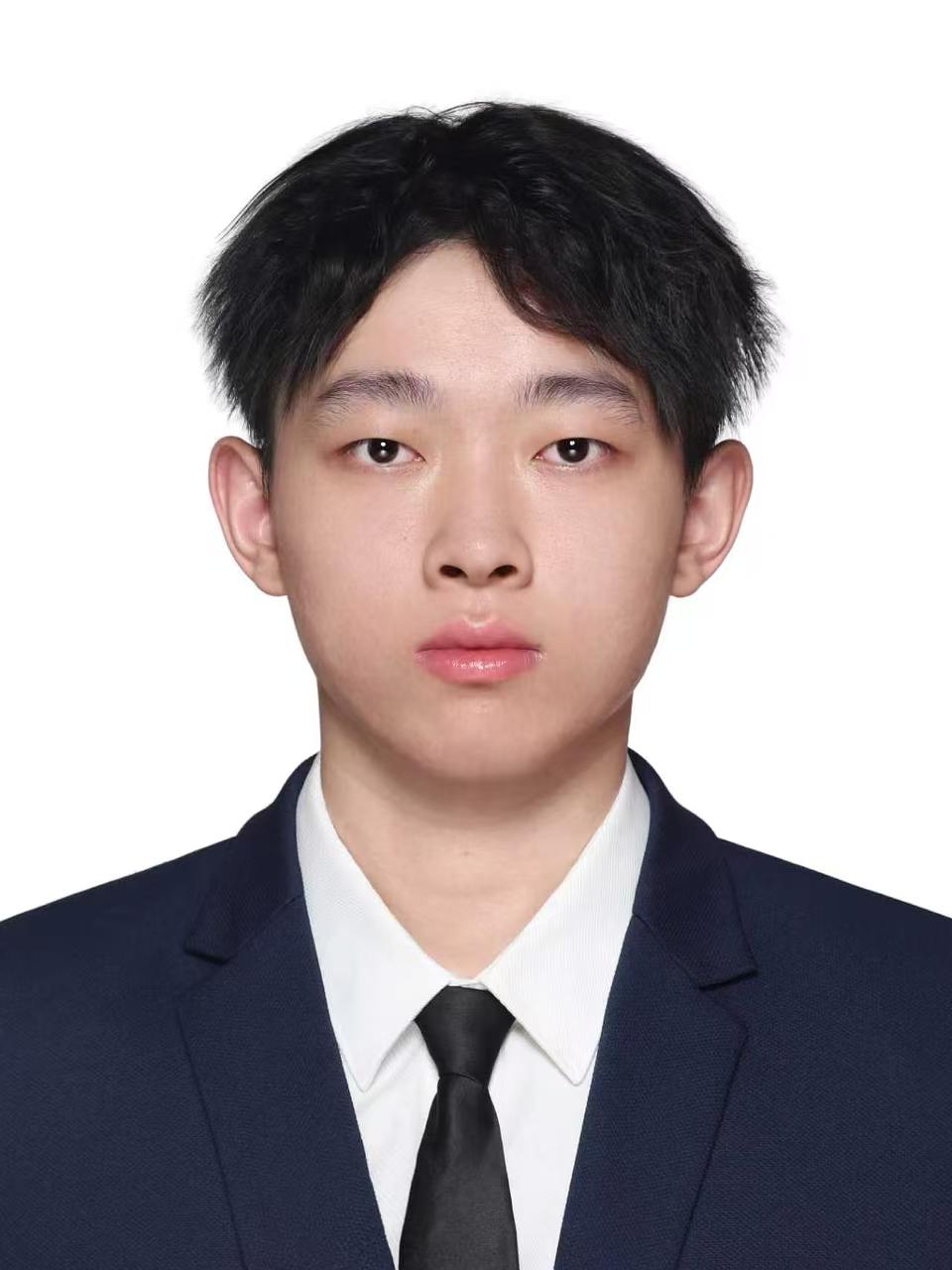}}]{Haotian Guo}
(Student Member, IEEE) received the B.Eng. degree in microelectromechanical systems engineering from Northwestern Polytechnical University, Xi'an, China, in 2026. He is currently pursuing the master's degree in mechanical engineering at the National University of Singapore. His research interests include tactile sensing, sensor design, and robotic perception.
\end{IEEEbiography}

\begin{IEEEbiography}[{\includegraphics[width=1in,height=1.25in,clip,keepaspectratio]{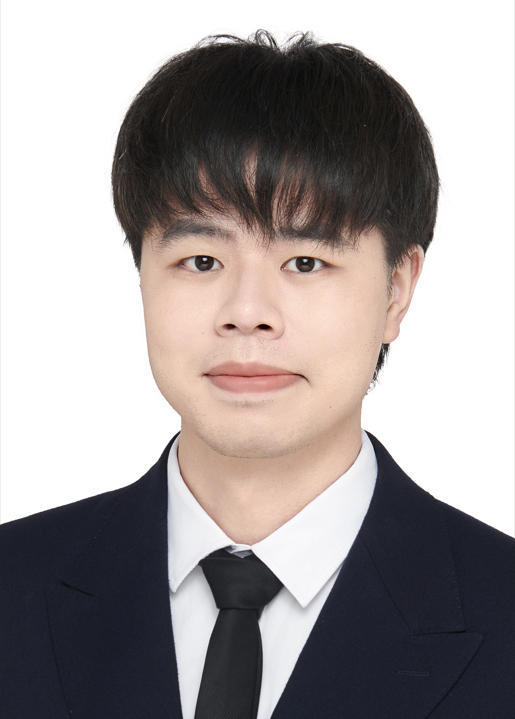}}]{Yu Feng}
(IEEE/ASME Member) received the B.Eng. degree in measurement, control, and automation from Harbin Engineering University, China, in 2020, and the Ph.D. degree in mechanical engineering from the City University of Hong Kong in 2024. He subsequently conducted postdoctoral research at the City University of Hong Kong from 2024 to 2026. He is currently a Research Fellow with the Department of Mechanical Engineering, National University of Singapore. His research interests include robotic perception, tactile sensors, and robotic manipulation.
\end{IEEEbiography}

\begin{IEEEbiography}[{\includegraphics[width=1in,height=1.25in,clip,keepaspectratio]{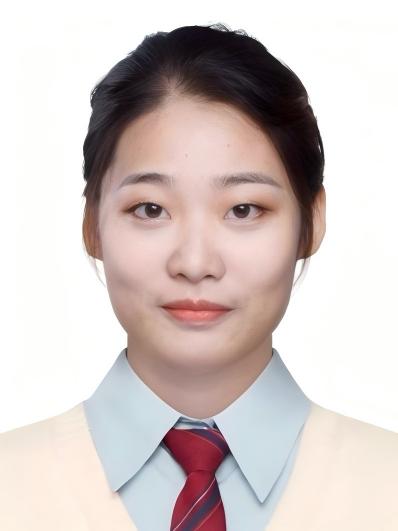}}]{Yutong Wang}
received the B.Eng. degree in automation from Harbin Institute of Technology, Weihai, China, in 2022, and the research master's (M.Phil.) degree in electronic information from the University of Chinese Academy of Sciences, Beijing, China, in 2025. Her research interests include robotics, deployable mechanisms, dexterous hand design, robot kinematics and dynamics, and robot control.
\end{IEEEbiography}

\begin{IEEEbiography}[{\includegraphics[width=1in,height=1.25in,clip,keepaspectratio]{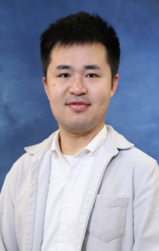}}]{Jianshu Zhou}
(IEEE/ASME Member) received the Ph.D. degree in mechanical engineering from the University of Hong Kong in 2020. He is currently an Assistant Professor with the Department of Mechanical Engineering, National University of Singapore. Before joining the National University of Singapore, he conducted research at the University of California, Berkeley, and the Chinese University of Hong Kong. His research interests include robotics and mechatronics, robotic grasping and manipulation, embodied intelligence and physical artificial intelligence, and multimodal sensing, data acquisition, and interaction. Dr. Zhou has received multiple Best Paper Awards and Best Paper Finalist Awards at international conferences, as well as the ASME DSCD Rising Star Award in 2025. He currently serves as an Associate Editor of IEEE Robotics and Automation Letters and a Technical Editor of IEEE/ASME Transactions on Mechatronics.
\end{IEEEbiography}

\end{document}